\documentclass[fleqn,10pt]{wlscirep}
\usepackage[utf8]{inputenc}
\usepackage[T1]{fontenc}
\usepackage{multirow}
\usepackage{booktabs}
\usepackage{tabularray}
\usepackage{ragged2e}
\usepackage{array}
\usepackage{graphicx}
\usepackage{subcaption}

\title{Assessing the Impact of Prompting Methods on ChatGPT's Mathematical Capabilities}

\author[1,*]{Yuhao Chen}
\author[1,*]{Chloe Wong}
\author[]{Hanwen Yang}
\author[]{Juan Aguenza}
\author[]{Sai Bhujangari}
\author[]{Benthan Vu}
\author[]{Xun Lei}
\author[]{Amisha Prasad}
\author[]{Manny Fluss}
\author[]{Eric Phuong}
\author[]{Minghao Liu}
\author[]{Raja Kumar}
\author[]{Vanshika Vats}
\author[]{James Davis}
\affil[]{University of California, Santa Cruz, 1156 High St, Santa Cruz, CA 95064, United States}
\affil[*]{ychen514@ucsc.edu, cwong771@ucsc.edu}

\begin{abstract}
This study critically evaluates the efficacy of prompting methods in enhancing the mathematical reasoning capability of large language models (LLMs). The investigation uses three prescriptive prompting methods - simple, persona, and conversational prompting - known for their effectiveness in enhancing the linguistic tasks of LLMs. We conduct this analysis on OpenAI’s LLM chatbot, ChatGPT-3.5, on extensive problem sets from the MATH, GSM8K, and MMLU datasets, encompassing a broad spectrum of mathematical challenges. A grading script adapted to each dataset is used to determine the effectiveness of these prompting interventions in enhancing the model’s mathematical analysis power. Contrary to expectations, our empirical analysis reveals that none of the investigated methods consistently improves over ChatGPT-3.5’s baseline performance, with some causing significant degradation. Our findings suggest that prompting strategies do not necessarily generalize to new domains, in this study failing to enhance mathematical performance.  

\end{abstract}

\begin{document}

\flushbottom
\maketitle
%
%
\thispagestyle{empty}
\section*{Introduction}
Large Language Models (LLM) have recently brought transformative advancements in the field of Natural Language Processing (NLP) \cite{ naveed2023comprehensive, devlin2018bert, radford2018improving}. Generative models such as Generative Pre-trained Transformer (GPT) \cite{radford2018improving}, Pathways Language Model (PaLM) \cite{chowdhery2023palm}, and Large Language Model Meta AI (LLaMA) \cite{touvron2023llama}, are trained with billions of parameters using terabytes of data, enabling them to have a deeper understanding of language and context, making them highly effective for a wide range of applications in NLP, robotics\cite{singh2023progprompt, song2023llm}, chat-bot \cite{kumar2023task} and medicine \cite{thirunavukarasu2023large }. 

ChatGPT \cite{chatgpt}, a variant of GPT designed to interact with humans in a natural conversational manner, demonstrates remarkable proficiency in generating coherent responses, and effectively addressing intricate challenges. A growing body of research has underscored the significance of prompt engineering in enhancing these models' accuracy for specialized tasks \cite{baidoo2023education, kocon2023chatgpt, giray2023prompt}. This technique has been instrumental in refining ChatGPT's performance across various fields, including medical engineering \cite{mesko2023prompt}, programming\cite{surameery2023use}, and software design \cite{white2023chatgpt}. There exist various forms of prompting techniques such as providing the chatbot a personality \cite{xu2022long}, guiding the context of the conversation relating to a specific field \cite{ge2023domain}, or giving it a sequential connection of ideas or concepts to guide response generation \cite{wei2022chain}. Although these methods have been shown to enhance language-related tasks, these improvements do not necessarily extend to areas where LLM inherently faces challenges, like complex mathematical problem-solving \cite{poola2023guiding}. 

In this work, we explore which prompting methods are most helpful in enhancing the mathematical problem-solving ability of large language models (LLM)\footnote{We use ChatGPT-3.5 in our work. In this manuscript, the terms ‘ChatGPT-3.5', ‘ChatGPT', and ‘chatbot' are used synonymously.}. We explore three broad categories - Simple, Persona, and Conversational - and ten specific prompting methods. We test these prompting methods on three different datasets, assessing accuracy on thousands of individual mathematical problems. Contrary to our expectations, we find that none of the evaluated prompting methods consistently enhance the mathematical problem-solving ability of ChatGPT.

A significant challenge in assessing the mathematical capabilities of LLM lies in the development of effective grading methodologies. Existing methods \cite{wei2022chain, frieder2023mathematical} predominantly rely on human evaluation to assess the accuracy and relevance of responses generated by these models. However, human evaluation poses limitations in terms of scalability. We create an automated grading script that is accurate and designed to minimize human intervention in the evaluation process. 



The primary contribution of this paper is evidence that prompt engineering strategies do not necessarily generalize to new domains. Our careful evaluation of ten specific prompting methods on three datasets found that none consistently improved the math problem-solving capability of ChatGPT.


\section*{Methods}
In our study, we use three datasets to evaluate three broad categories of prompting with ten specific prompts.  An automated grading script called GPT-Grader is used to compute accuracy. 

\begin{table}[]
\centering
\begin{tabular}{@{}|cl|l|@{}}
\toprule
\multicolumn{2}{|c|}{\textbf{Prompting Methods}} &
  \multicolumn{1}{c|}{\textit{\textbf{Prompts}}} \\ \midrule
\multicolumn{1}{|c|}{\multirow{4}{*}{\begin{tabular}[c]{@{}c@{}}Simple \\ Prompting\end{tabular}}} &
  Topic Prompt &
  \textit{This is a \textless{}Math/Algebra/Geometry\textgreater question} \\ \cmidrule(l){2-3} 
\multicolumn{1}{|c|}{} &
  "Difficult" Prompt &
  \textit{This is a difficult question} \\ \cmidrule(l){2-3} 
\multicolumn{1}{|c|}{} &
  "Easy" Prompt &
  \textit{This is an easy question} \\ \cmidrule(l){2-3} 
\multicolumn{1}{|c|}{} &
  Calculation Prompt &
  \textit{You are an accurate calculator, please calculate the provided equation} \\ \midrule
\multicolumn{1}{|c|}{\multirow{3}{*}{\begin{tabular}[c]{@{}c@{}}Persona \\ Prompting\end{tabular}}} &
  High Confidence &
  \textit{\begin{tabular}[c]{@{}l@{}}You are a person with very high confidence in answering math questions, respond \\ to the following question, I think you can do it!\end{tabular}} \\ \cmidrule(l){2-3} 
\multicolumn{1}{|c|}{} &
  Low Confidence &
  \textit{\begin{tabular}[c]{@{}l@{}}You are a person with very low confidence in answering math questions, respond \\ to the following question, but I doubt you will be able to answer it!\end{tabular}} \\ \cmidrule(l){2-3} 
\multicolumn{1}{|c|}{} &
  No Explanation &
  \textit{\begin{tabular}[c]{@{}l@{}}You are a person who gives no explanation to any of your steps, respond to the \\ following question\end{tabular}} \\ \midrule
\multicolumn{1}{|c|}{\multirow{3}{*}{\begin{tabular}[c]{@{}c@{}}Conversational \\ Prompting\end{tabular}}} &
  Casual Conversation &
  \textit{\begin{tabular}[c]{@{}l@{}}P1: Hello, how’s your day GPT?\\ P2: I heard you helped a lot of people solve different problems, you are so great!\end{tabular}} \\ \cmidrule(l){2-3} 
\multicolumn{1}{|c|}{} &
  Math Conversation &
  \textit{\begin{tabular}[c]{@{}l@{}}P1: What do you know about the Riemann?\\ P2: What do you know about math, do you think math is hard?\end{tabular}} \\ \cmidrule(l){2-3} 
\multicolumn{1}{|c|}{} &
  Physics Conversation &
  \textit{\begin{tabular}[c]{@{}l@{}}P1: What do you know about gravity?\\ P2: Can you list some famous physics experiments?\end{tabular}} \\ \bottomrule
\end{tabular}
\caption{Ten specific \emph{prompts} were used to test Simple, Persona and Conversational prompting methods. Here, \emph{P1} and \emph{P2} refer to two-shot prompts used to initiate chats in conversational prompting.}
\label{table:prompts}
\end{table}

\subsection*{Datasets}
We use existing datasets - MATH\cite{hendrycksmath2021}, GSM8K\cite{cobbe2021gsm8k}, and MMLU\cite{hendrycksmmlu2021, hendrycks2021ethics} - to evaluate ChatGPT’s mathematical capabilities. The MATH dataset consists of 12,500 mathematics problems encompassing various types, including linear algebra, calculus, geometry, number theory, and statistics. Each problem comes with a full step-by-step solution as the ground truth. The GSM8K dataset comprises 8.5K high-quality, linguistically diverse grade school math word problems having natural language as well as a final numeric solution. The MMLU dataset, or Measuring Massive Multitask Language Understanding, covers 57 tasks, including elementary mathematics, United States history, computer science, and law. For our study, we only test ChatGPT on math-related categories, including abstract algebra, college mathematics, elementary mathematics, and high school mathematics. Notably, MMLU differs from the other two datasets in that it employs multiple-choice questions and expects one correct choice instead of short answers.

\subsection*{Prompting Methods}
\begin{figure}
    \centering
    \begin{subfigure}[b]{0.33\textwidth}
        \includegraphics[width=\textwidth]{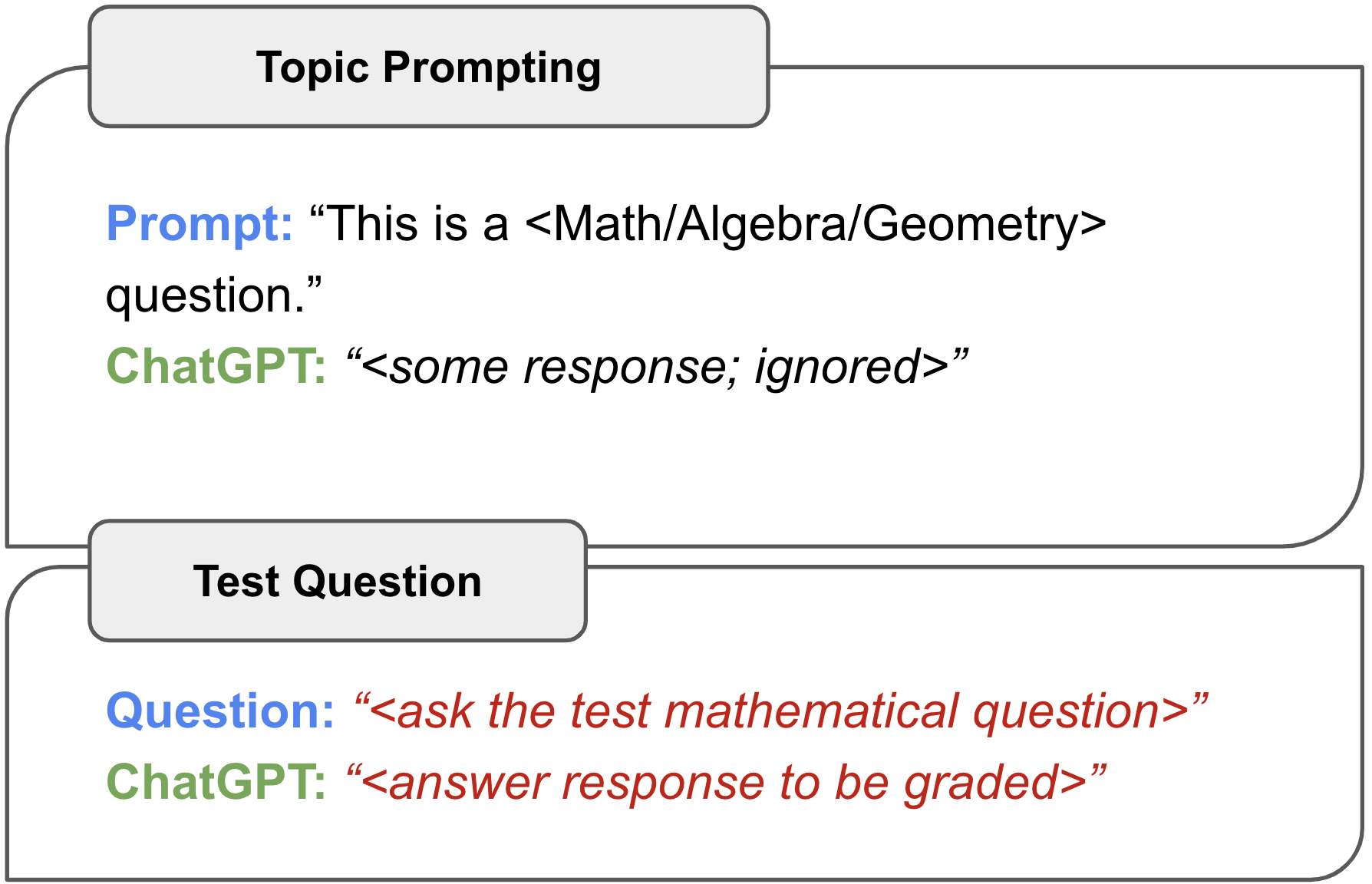}
        \caption{}
        \label{fig:simple}
    \end{subfigure}
    \begin{subfigure}[b]{0.33\textwidth}
        \includegraphics[width=\textwidth]{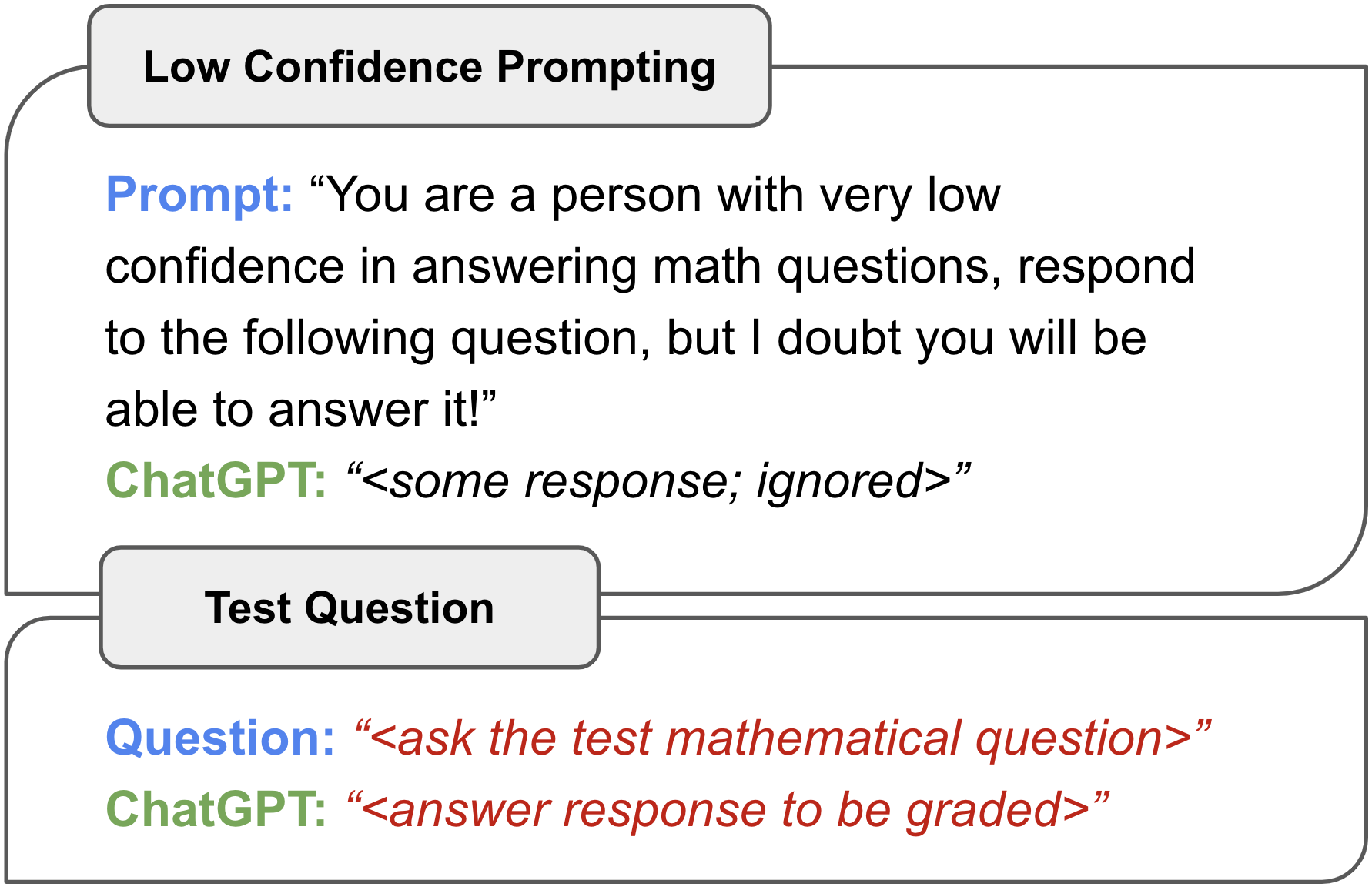}
        \caption{}
        \label{fig:persona}
    \end{subfigure}
    \begin{subfigure}[b]{0.33\textwidth}
        \includegraphics[width=\textwidth]{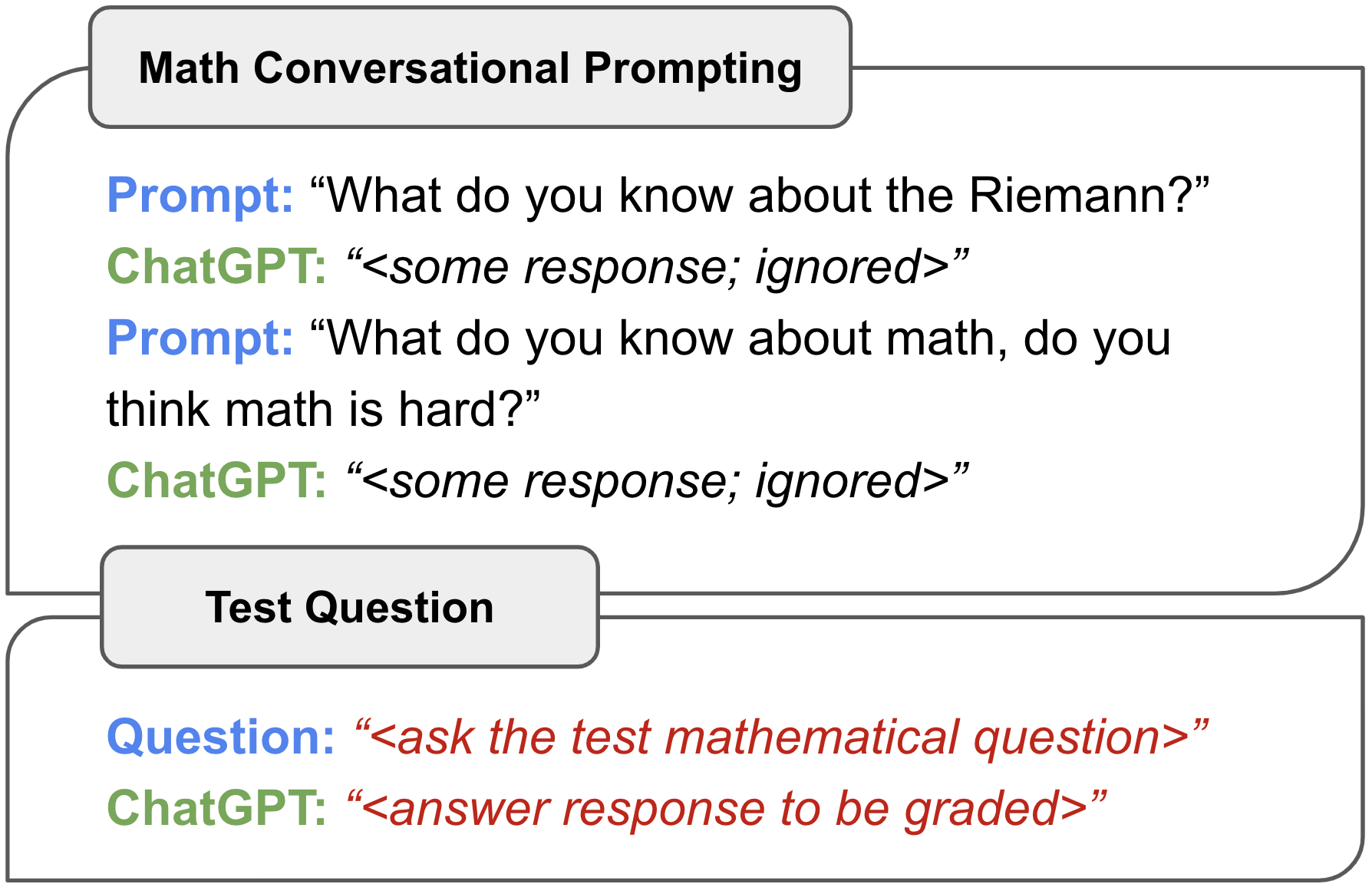}
        \caption{}
        \label{fig:conversation}
    \end{subfigure}
    \vspace{-20pt}
    \caption{\textbf{Illustration of prompting workflow:} The figure presents an example each from (a) simple, (b) persona, and (c) conversational prompting methods. We initiate the conversation with ChatGPT using the defined prompts. We evaluate only the response to the subsequently asked mathematical question.}
    \label{fig:cot_prompts}
\end{figure}

Prompting is simply giving an LLM inputs for the model to respond to prior to the actual question of interest\cite{giray2023prompt}. 
We assess ChatGPT-3.5's proficiency in mathematical reasoning by analyzing its accuracy in responding to math questions presented after a range of prompting techniques, including simple, persona, and conversational prompting. We design specific prompts within each category. The specific prompts used in our study are shown in Table \ref{table:prompts} and an illustration of prompting workflow is provided in Fig. \ref{fig:cot_prompts}.

\subsubsection*{Simple Prompting}
This approach includes prompting methods used to bias the chatbot at the start of the conversation. Our simple prompting approach analyzes topic-based, difficulty-based, and calculation-based prompting methods. In topic prompting, we provide topic-specific prompts in the hope of nudging ChatGPT towards the right context. We hypothesize that giving the chatbot a hint regarding the type of mathematical question we are about to ask  (Math/Algebra/Geometry) helps it render accurate answers. With difficulty-based prompting we test whether ChatGPT’s ability to respond correctly varies with respect to how easy or difficult we have told it the question is. This is done to see if biasing the confidence level from the start would affect ChatGPT’s efficacy in giving the right answer. In calculation prompting, ChatGPT is directed to behave as an accurate calculator and calculate the given mathematical problem.

\subsubsection*{Persona Prompting}
Persona prompting gives an LLM an identity or personality to adopt while generating the responses\cite{liu2022personabased}. Joshi et al.\cite{Joshi2023PersonasAA} hypothesize that large models tend to generate truthful responses if they adopt a credible persona. A high-confidence persona is created in hopes of ChatGPT mimicking someone who performs above average. Online sources are from a wide array of expertise and backgrounds. The hope is that the LLM will seek credible resources after pre-affirming its decision-making. We also test a low-confidence persona and see if the chatbot's decision-making dwindles. 

In initial tests we noted that LLMs often produce flawed reasoning and ‘hallucinations’\cite{zhang2023hallucination}, where it delivers varied responses with high confidence despite lower accuracy rates. We thus test whether instructing ChatGPT to condense its explanations would be beneficial. We ask the chatbot to assume a persona that does not give explanations.

\subsubsection*{Conversational Prompting}
We test conversational prompting by engaging in a conversation with the chatbot to sequentially guide it by providing some context of our expected results, taking inspiration from context-faithful prompting\cite{zhou2023context}.  Our motivation for prompting ChatGPT with initial two-shot conversations is to test whether it performs better when guided with sequential links of thoughts toward the right field of interest. The casual conversation starts with setting a casual tone for the chatbot. This  choice is aimed at examining the bot's response behavior in a relaxed, conversational setting. For the more academically inclined conversations, we test mathematical and physics-focused conversations. In the former, the conversation encompasses a  mathematical field and investigates whether ChatGPT is able to take the hints. The prompts are structured to initiate a discussion about math with the aim of pre-conditioning the internal state of the chatbot prior to asking the mathematical question. The physics-focused conversation provides a pre-condition distinct from the mathematics questions that we will subsequently pose to the bot.


\subsection*{Grading Methods}
\begin{figure}
    \centering
    \begin{subfigure}[b]{0.33\textwidth}
        \includegraphics[width=\textwidth]{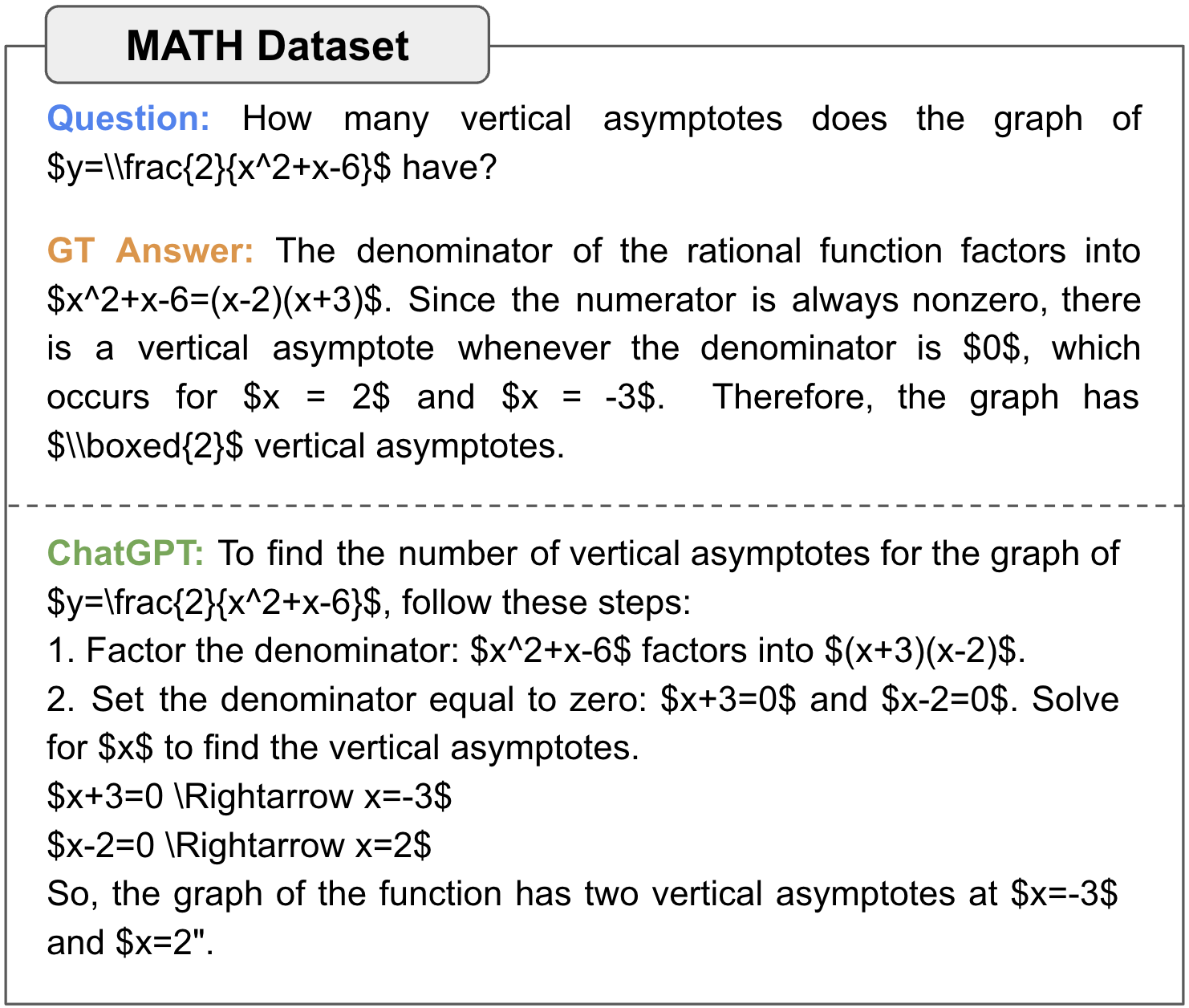}
        \caption{}
        \label{fig:MATH}
    \end{subfigure}
    \begin{subfigure}[b]{0.33\textwidth}
        \includegraphics[width=\textwidth]{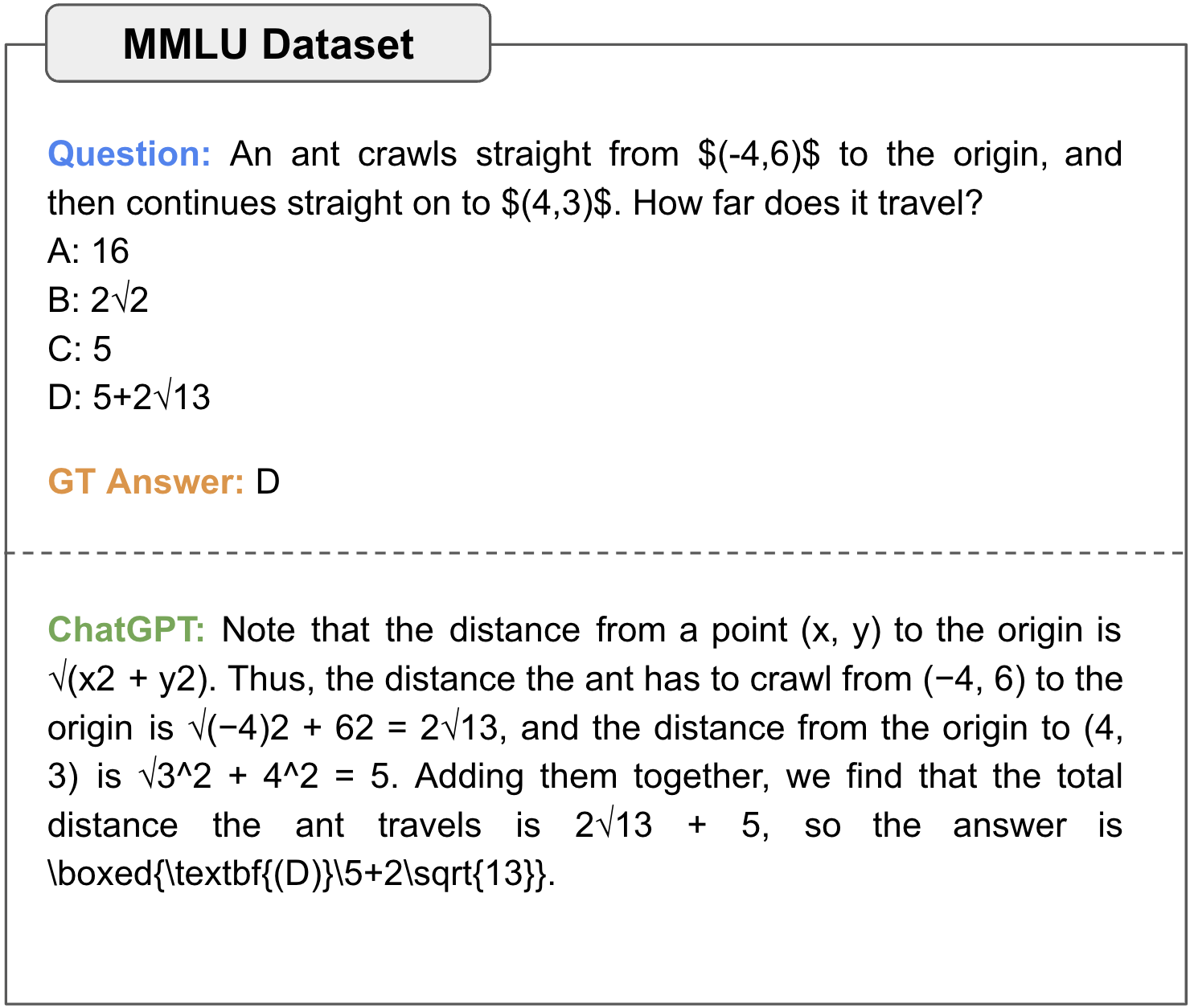}
        \caption{}
        \label{fig:MMLU}
    \end{subfigure}
    \begin{subfigure}[b]{0.33\textwidth}
        \includegraphics[width=\textwidth]{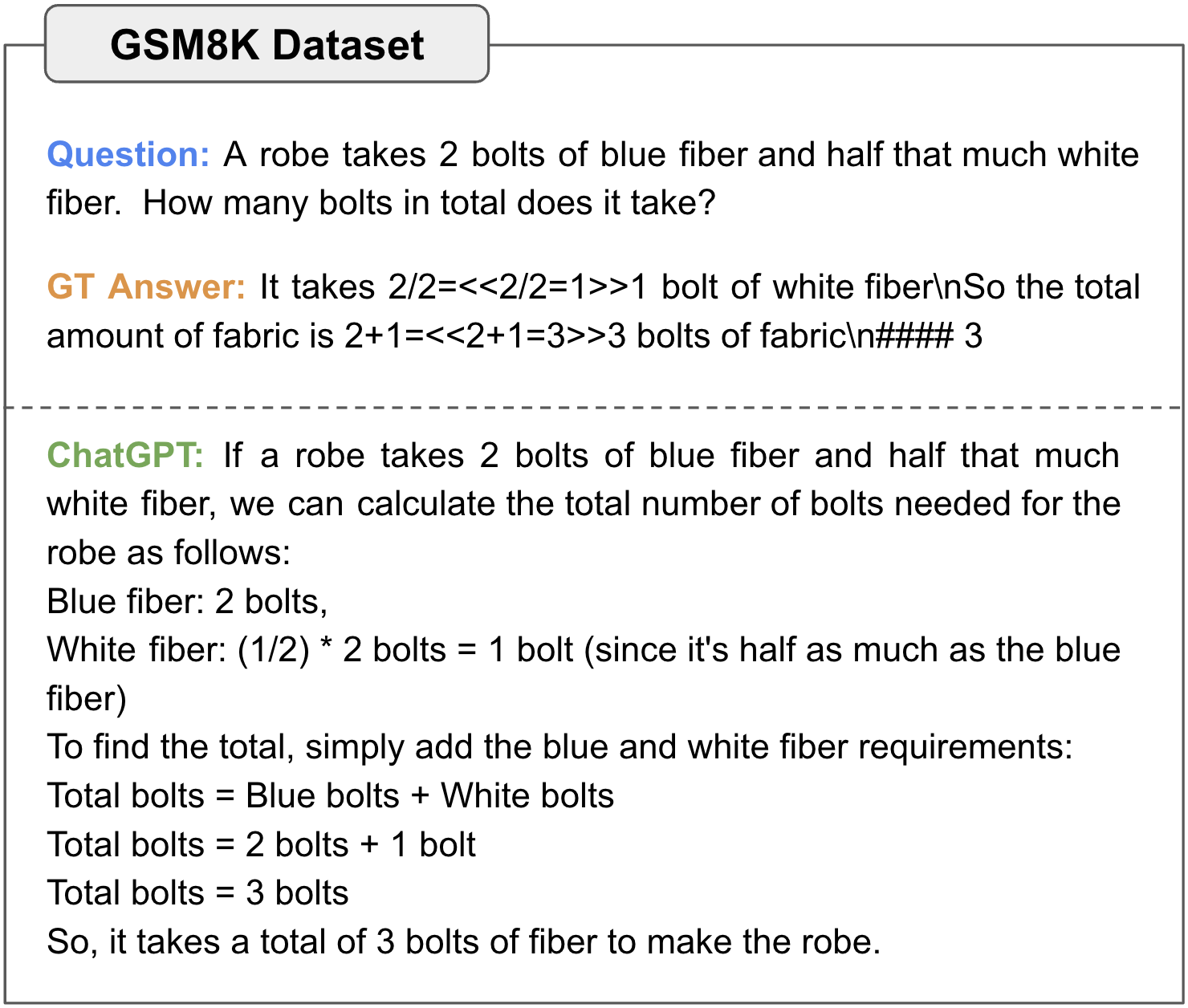}
        \caption{}
        \label{fig:GSM8K}
    \end{subfigure}
    \vspace{-20pt}
    \caption{\textbf{Examples of question-answer pairs in each datasets:} The figure presents an example each from the (a) MATH, (b) MMLU, and (c) GSM8K datasets, along with their dataset provided ground truth answers (GT). Below each example is ChatGPT's response for comparison. Notice that ChatGPT's responses are neither straightforward nor completely aligned with the ground truth format. We thus need specially designed grading methods for evaluation of accuracy.}
    \label{fig:datasets}
\end{figure}

We develop a grading script, GPT-Grader, to auto-grade ChatGPT’s response. Neither the datasets nor ChatGPT consistently provide answers in a format which allows easy direct comparison. An example from each dataset illustrating the difficulty is shown in Fig.~\ref{fig:datasets}. GPT-Grader uses a grading method customized for the ground truth format of each dataset. 

Given the variability of the response format in the MATH dataset, we process the resulting responses to make them comparable. For example, the final response of 0.25 and ¼ mean the same thing.  After obtaining ChatGPT’s response to a problem from the MATH dataset, the GPT-Grader script first looks for a direct match between the solutions. In the absence of a match, it next calculates the numerical value of each solution and compares these answers. If a match still does not exist, GPT-Grader searches the 
 entire ChatGPT response for the ground truth solution. This extraction is done with the help of RegEx and string slicing. If the solution is identified anywhere in the response, the script verifies that the match is embedded within a mathematical equation. 


The MMLU dataset has a multiple-choice format for answers. ChatGPT does not provide a simple response, instead embedding the answer in a paragraph of text. Since the valid answers of A,B,C,D also appear in unrelated text strings, simple matching does not work well. Instead, GPT-Grader uses ChatGPT itself to parse the response. The script resets the history of ChatGPT and resubmits the solution to it, instructing the model to identify the letter answer within the solution and format it as either [A], [B], [C], or [D]. Ideally, for the example in Fig.~\ref{fig:MMLU}, a valid response would be [D] or D. Resetting the history is important so that any previous context does not introduce bias into the new comparison. This step is repeated until a valid response is provided or until eight unsuccessful attempts. Finally, once a response is obtained, the script compares it with the actual ground truth answer from the dataset. If the response matches the ground truth, it is marked as correct; otherwise, it is incorrect. 


The grading approach for the GSM8K dataset also uses ChatGPT to help grade its own responses. Instead of requesting ChatGPT to format the solution as [A], [B], [C], or [D] in the second step, we provide the ground truth solutions and ChatGPT's solution to the model as a new query. We ask ChatGPT to verify the accuracy of the answer, parsing this new response using a  methodology similar to MMLU evaluation. 

To validate the performance of our grading methods, we conduct manual evaluations on a random sample of responses from each method. We find an error rate of 1.35\% on the MATH dataset (16 errors out of 1187 result samples), 1.54\% on GSM8K (131 errors out of 8500 result samples), and 0.47\% on MMLU (4 errors out of 848 result samples). We consider the average error rate of less than 2\% sufficient for this study. 
\section*{Results and Discussion}
\begin{figure}[ht!]
\centering
\includegraphics[width=0.9\linewidth]{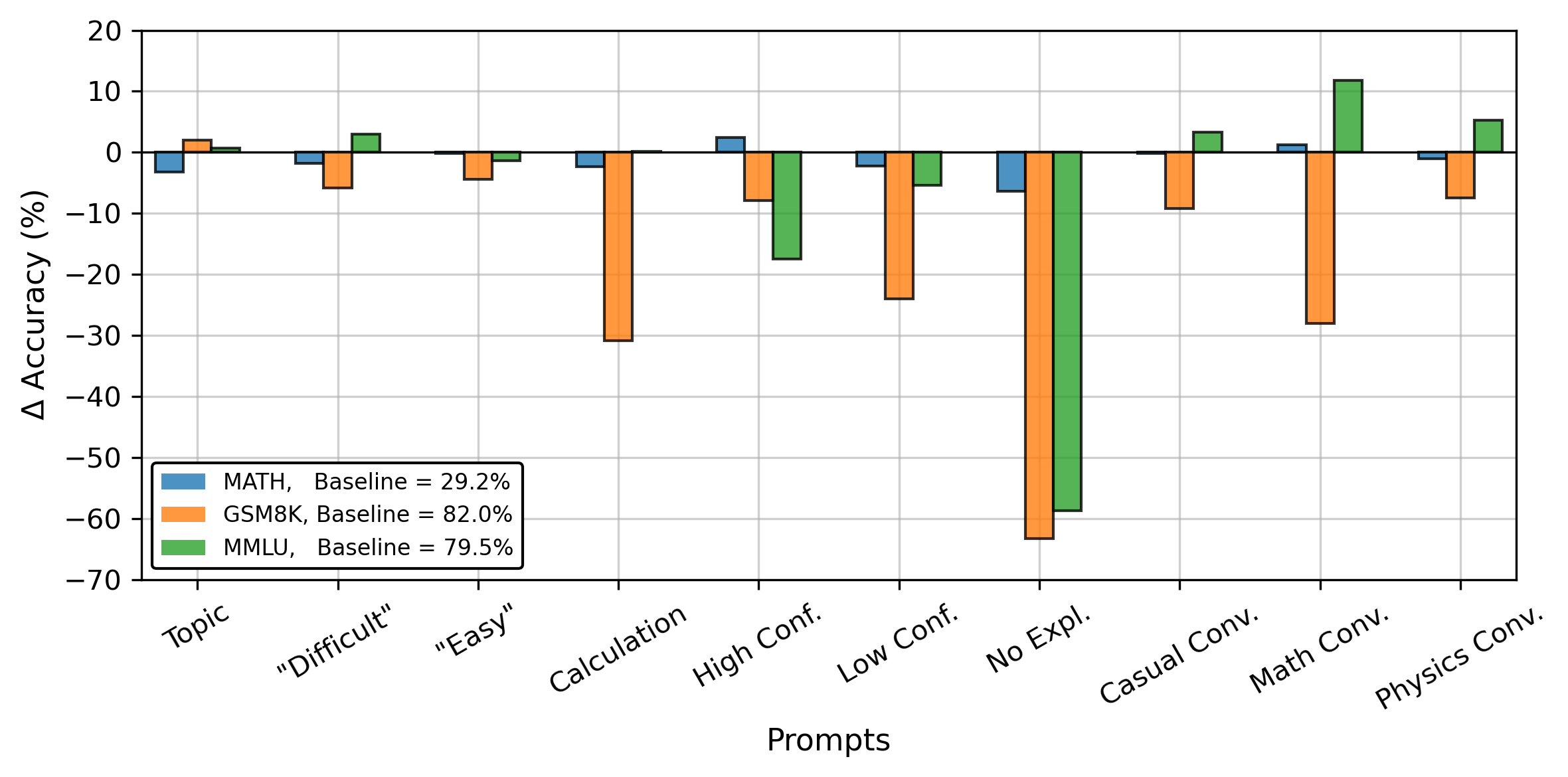}
\vspace{-10pt}
\caption{\textbf{Visualization of change in accuracy:}  Baseline accuracy for each dataset is normalized to \emph{0 (zero)} on the y-axis. Bars denote \emph{$\Delta$ Accuracy}, improvement or deterioration in the performance when using prompts with respect to the baseline. Notice that the only significant improvement from prompting was "Math Conversation" on the MMLU dataset. However this prompting method showed a significant degradation of performance on the GSM8K dataset. We conclude that none of the evaluated prompting methods provide generalizable gains on mathematical performance.}
\label{fig:delta_result}
\end{figure}

To examine whether the given prompting techniques are effective, we assess the accuracy on the MATH, GSM8K, and MMLU datasets.   Each prompting method \emph{(<prompt, input, output>)} is compared with the baseline \emph{(<input, output>)}, with results provided in Table \ref{table:results}. A visualization of the change in accuracy is provided in Fig. \ref{fig:delta_result}. We do not observe consistently improved outcomes for any of the prompting methods tested. 

Among the simple prompting methods, topic prompting shows a slight improvement on the GSM8K and MMLU datasets, but a decrease in accuracy on the MATH dataset. Difficulty prompting does not show any remarkable improvement compared to the baseline results. Asking ChatGPT to behave as an accurate calculator also does not show any enhancement. 
\begin{table}[]
\centering
\begin{tabular}{@{}|cl|rrrrrr|@{}}
\toprule
\multicolumn{1}{|c|}{\multirow{2}{*}{\textbf{\begin{tabular}[c]{@{}c@{}}Prompting \\ Methods\end{tabular}}}} &
  \multicolumn{1}{c|}{\multirow{2}{*}{\textbf{Prompts}}} &
  \multicolumn{6}{c|}{\textbf{Accuracy (\%)}} \\ \cmidrule(l){3-8} 
\multicolumn{1}{|c|}{} &
  \multicolumn{1}{c|}{} &
  \multicolumn{1}{c|}{\textbf{MATH}} &
  \multicolumn{1}{c|}{\textbf{$\Delta_{\text{MATH}}$}} &
  \multicolumn{1}{c|}{\textbf{GSM8K}} &
  \multicolumn{1}{c|}{\textbf{$\Delta_{\text{GSM8K}}$}} &
  \multicolumn{1}{c|}{\textbf{MMLU}} &
  \multicolumn{1}{c|}{\textbf{$\Delta_{\text{MMLU}}$}} \\ \midrule
\multicolumn{2}{|c|}{Baseline} &
  \multicolumn{1}{r|}{29.2} &
  \multicolumn{1}{r|}{0} &
  \multicolumn{1}{r|}{82.0} &
  \multicolumn{1}{r|}{0} &
  \multicolumn{1}{r|}{79.5} &
  0 \\ \midrule
\multicolumn{1}{|c|}{\multirow{4}{*}{\begin{tabular}[c]{@{}c@{}}Simple \\ Prompting\end{tabular}}} &
  Topic Prompt &
  \multicolumn{1}{r|}{26.0} &
  \multicolumn{1}{r|}{-3.2} &
  \multicolumn{1}{r|}{84.0} &
  \multicolumn{1}{r|}{+2.0} &
  \multicolumn{1}{r|}{80.2} &
  +0.7 \\
\multicolumn{1}{|c|}{} &
  "Difficult" Prompt &
  \multicolumn{1}{r|}{27.4} &
  \multicolumn{1}{r|}{-1.8} &
  \multicolumn{1}{r|}{76.2} &
  \multicolumn{1}{r|}{-5.8} &
  \multicolumn{1}{r|}{82.5} &
  +3.0 \\
\multicolumn{1}{|c|}{} &
  "Easy" Prompt &
  \multicolumn{1}{r|}{29.0} &
  \multicolumn{1}{r|}{-0.2} &
  \multicolumn{1}{r|}{77.6} &
  \multicolumn{1}{r|}{-4.4} &
  \multicolumn{1}{r|}{78.1} &
  -1.4 \\
\multicolumn{1}{|c|}{} &
  Calculation Prompt &
  \multicolumn{1}{r|}{26.8} &
  \multicolumn{1}{r|}{-2.4} &
  \multicolumn{1}{r|}{51.2} &
  \multicolumn{1}{r|}{-30.8} &
  \multicolumn{1}{r|}{79.7} &
  +0.2 \\ \midrule
\multicolumn{1}{|c|}{\multirow{3}{*}{\begin{tabular}[c]{@{}c@{}}Persona \\ Prompting\end{tabular}}} &
  High Confidence &
  \multicolumn{1}{r|}{31.6} &
  \multicolumn{1}{r|}{+2.4} &
  \multicolumn{1}{r|}{74.1} &
  \multicolumn{1}{r|}{-7.9} &
  \multicolumn{1}{r|}{62.0} &
  -17.5 \\
\multicolumn{1}{|c|}{} &
  Low Confidence &
  \multicolumn{1}{r|}{27.0} &
  \multicolumn{1}{r|}{-2.2} &
  \multicolumn{1}{r|}{58.0} &
  \multicolumn{1}{r|}{-24.0} &
  \multicolumn{1}{r|}{74.1} &
  -5.4 \\
\multicolumn{1}{|c|}{} &
  No Explanation &
  \multicolumn{1}{r|}{22.8} &
  \multicolumn{1}{r|}{-6.4} &
  \multicolumn{1}{r|}{18.8} &
  \multicolumn{1}{r|}{-63.2} &
  \multicolumn{1}{r|}{20.8} &
  -58.7 \\ \midrule
\multicolumn{1}{|c|}{\multirow{3}{*}{\begin{tabular}[c]{@{}c@{}}Conversational \\ Prompting\end{tabular}}} &
  Casual Conversation &
  \multicolumn{1}{r|}{29.0} &
  \multicolumn{1}{r|}{-0.2} &
  \multicolumn{1}{r|}{72.8} &
  \multicolumn{1}{r|}{-9.2} &
  \multicolumn{1}{r|}{82.8} &
  +3.3 \\
\multicolumn{1}{|c|}{} &
  Math Conversation &
  \multicolumn{1}{r|}{30.4} &
  \multicolumn{1}{r|}{+1.2} &
  \multicolumn{1}{r|}{54.0} &
  \multicolumn{1}{r|}{-28.0} &
  \multicolumn{1}{r|}{91.3} &
  +11.8 \\
\multicolumn{1}{|c|}{} &
  Physics Conversation &
  \multicolumn{1}{r|}{28.2} &
  \multicolumn{1}{r|}{-1.0} &
  \multicolumn{1}{r|}{74.5} &
  \multicolumn{1}{r|}{-7.5} &
  \multicolumn{1}{r|}{84.8} &
  +5.3 \\ \bottomrule
\end{tabular}
\caption{\textbf{Performance of prompting methods:} The accuracy of each prompt is evaluated on the MATH, GSM8K, and MMLU datasets. Performance change from the baseline ($\Delta_{\text{Dataset}}$) is also provided. Notice that results are inconsistent across datasets, with none of the prompting methods providing a reliable improvement in accuracy.}
\label{table:results}
\end{table}

In persona prompting, giving a low-confidence persona to the bot consistently results in poor performance. The high-confidence persona does show improvement on the MATH dataset, but degradation on the other tests. The no-explanation persona was hypothesized to improve performance by reducing hallucinations\cite{zhang2023hallucination} and false reasoning patterns. Unfortunately, the results consistently show a dramatic drop in accuracy from the baseline. We conclude that bypassing step-by-step reasoning can cause the large model to skip important workflow, resulting in wrong answers.

Analyzing the results from the guided sequential conversational prompting experiments reveals varied outcomes across the datasets. The MMLU dataset exhibits a significant boost in accuracy when the chatbot was preconditioned to converse about math. However, the GSM8K dataset shows a significant reduction in the accuracy. If we had only tested the MMLU dataset, we might have concluded that this prompting method enhances the LLM's mathematical capability. Testing prompting strategies on multiple datasets appears to be important for verifying the generalizability of results. Interestingly the out-of-domain casual and physics-based conversation prompts also led to improvements on the MMLU dataset.

Past researches\cite{xu2022long,Joshi2023PersonasAA} indicate a beneficial influence of prompting on language tasks. However, our combined results on mathematical tasks do not find any of the tested prompting methods to provide consistent improvement. There are, of course, limitations to this finding. We test only one specific LLM, GPT-3.5, and other LLMs\cite{chowdhery2023palm,touvron2023llama} might behave differently. Similarly, we tested only ten specific prompts on only three specific datasets. Nevertheless, we believe these results show that prompting strategies do not necessarily generalize from one domain to another or even from one dataset to another within the same domain.

\section*{Conclusion}
We test three prompting methods using ten specific prompts to determine if prompting enhances ChatGPT performance on mathematical tasks.  Contrary to initial expectations, we find that none of the tested methods consistently improve results, sometimes improving accuracy on one dataset while degrading accuracy on another. Our findings suggest that prompting strategies do not necessarily generalize to new datasets and domains.

\bibliography{sample}



\section*{Author contributions statement}
Y.C., H.Y., M.L., and J.D. designed the study; J.A., B.V., S.B., C.W., X.L., A.P., M.F., E.P., and Y.C. ran experiments and evaluations; Y.C, C.W., H.Y., and A.P. wrote the first draft of the manuscript; V.V. developed tables and visual representations; V.V and R.K. reviewed, refined, and edited the final draft.

\end{document}